\documentclass[11pt,a4paper]{article}
\usepackage[hyperref]{naaclhlt2018}
\usepackage{times}
\usepackage{latexsym}
\usepackage{amsmath}
\usepackage{amssymb}
\usepackage{booktabs}
\usepackage{graphicx}
\usepackage{array}
\usepackage{xcolor} 

\newcolumntype{L}{>{\centering\arraybackslash}m{3cm}}

\usepackage{url}
\usepackage{etoolbox}
\makeatletter
\patchcmd\@combinedblfloats{\box\@outputbox}{\unvbox\@outputbox}{}{%
   \errmessage{\noexpand\@combinedblfloats could not be patched}%
}%
 \makeatother

\newcommand{\eat}[1]{}

\aclfinalcopy 

\newcommand*{\affaddr}[1]{#1} 
\newcommand*{\affmark}[1][*]{\textsuperscript{#1}}
\newcommand*{\email}[1]{\texttt{#1}}

\title{Controlling Decoding for More Abstractive Summaries \\ with Copy-Based Networks}

\author{%
Noah Weber\thanks{*These authors contributed equally to this work.} \affmark[1], Leena Shekhar\footnotemark[1] \affmark[1], Niranjan Balasubramanian\affmark[1], Kyunghyun Cho\affmark[2] \\
\affaddr{\affmark[1]Stony Brook University, NY}\\
\email{\{nwweber, lshekhar, niranjan\}@cs.stonybrook.edu}\\
\affaddr{\affmark[2]New York University, NY}\\
\email{Kyunghyun.cho@nyu.edu} %
}

\date{}

\begin{document}
\maketitle
\begin{abstract}
Attention-based neural abstractive summarization systems equipped with copy mechanisms have shown promising results. Despite this success, it has been noticed that such a system generates a summary by mostly, if not entirely, copying over phrases, sentences, and sometimes multiple consecutive sentences from an input paragraph, effectively performing extractive summarization. In this paper, we verify this behavior using the latest neural abstractive summarization system - a pointer-generator network~\citep{see2017get}. We propose a simple baseline method that allows us to control the amount of copying without retraining. Experiments indicate that the method provides a strong baseline for abstractive systems looking to obtain high ROUGE scores while minimizing overlap with the source article, substantially reducing the n-gram overlap with the original article while keeping within 2 points of the original model's ROUGE score.
 \end{abstract}

\section{Introduction}

Automatic abstractive summarization has seen a renewed interest in recent years~\citep{rush2015neural,Nallapati2016AbstractiveTS,see2017get} building on attention-based encoder-decoder models originally proposed for neural machine translation~\citep{bahdanau2014neural}. Recent approaches rely on encoder-decoder formulations augmented with copy-mechanisms to produce abstractive summaries. The encoder-decoder allows the model to generate new words that are not part of the input article, while the copy-mechanism allows the model to copy over important details from the input even if these symbols are rare in the training corpus overall. \citet{see2017get} and \citet{paulus2017deep} use a pointer-generator model which produces a summary using an interpolation of generation and copying probabilities. The interpolation is controlled by a mixture coefficient that is predicted by the model at each time step.

Even though the pointer-generator mechanism, in theory, enables the model to interpolate between extractive (copying) and abstractive (generating) modes, in practice the extractive mode dominates. \citet{see2017get} for instance reported that ``at test time, $\left[ \text{the conditional distribution is} \right]$ heavily skewed towards copying''. This is also evident from the examples from the state-of-the-art system presented in Table 3 of \citet{paulus2017deep}.

We carefully confirm this behavior using the neural abstractive summarization system by \citet{see2017get}. We consider the $n$-gram overlap between the input paragraph and generated summary and observe extremely high overlaps across varying $n$'s (from 2 up to 25). When the coverage penalty, which was found to improve the summarization quality in \citet{see2017get}, was introduced, these overlaps further increased. On the other hand, ground-truth summaries have almost no overlaps. This clearly suggests that the neural abstractive summarization system largely performs extractive summarization. 

We introduce a simple modification to beam search to promote abstractive modes during decoding. In particular, for each hypothesis we track the mixture coefficients that are used to combine the copying and generating probabilities at each time step. As \citet{see2017get} report, during test time, 
the mixture coefficients are often significantly low, which predominantly favors copying. To counter this effect, we introduce an additional term to the beam score, which
penalizes a hypothesis whose average mixture coefficient deviates from a expected mixture, predefined target. By setting the target appropriately, it allows us to control the level of abstractiveness at the decoding time. We empirically confirm that we can control the abstractiveness while \textit{largely maintaining the quality of summary}, measured in terms of ROUGE~\citep{lin2004rouge} and METEOR~\citep{lavie2009meteor}, without having to retrain a system. The relative simplicity and performance of the method makes it a strong baseline for future abstractive summarization systems looking to solve the copying problem during training.

\section{Neural Abstractive Summarization and Copy-Controlled Decoding}

\subsection{Neural Abstractive Summarization}

In this paper, we use the pointer-generator network, proposed by \citet{see2017get}, as a target abstractive summarization system. The pointer-generator network is an extension of an earlier neural attention model for abstractive sentence summarization by \citet{rush2015neural} that incorporates the copy mechanism by \cite{gu2016incorporating}. 
Since our focus is on the copying behavior, we summarize the decoder component of the pointer-generator network here. We refer the readers to \citet{see2017get} for other details.

At each time step $t$, the decoder computes three quantities: (i) a copy distribution over the source symbols in the input $p_{\text{copy}}$, (ii) a generating distribution $p_{\text{gen}}$ defined over a predefined vocabulary (all the symbols in the training data), and (iii) a mixture co-efficient, $m_t$, that is used to combine the copy and generating distributions.

The decoder computes the copy distribution at time $t$ via the attention weights $\alpha_i$'s defined over the encoded representations $h_i$'s of the corresponding source symbols $x_i$'s. 
Since these weights are non-negative and sum to one, we can treat them as the output probability distribution defined over the source symbols, which we refer to as the copy distribution $p_{\text{copy}}$.
Then, the decoder computes the generating distribution $p_{\text{gen}}$ over the entire training vocabulary based on the context vector, $h_t^*=\sum_{i=1}^{|X|} \alpha_i h_i$, and the decoder's state $s_t$. These two distributions are then mixed based on a coefficient $m_t \in \left[0, 1\right]$, which is also computed based on the context vector $h_t^*$, the decoder's hidden state $s_t$ and the previously decoded symbol $\hat{y}_{t-1}$. The final output distribution is given by
\begin{align*}
p(w) =& m_t p_{\text{gen}}(w)+ (1-m_t) \sum_{i=1}^{|X|} \mathbb{I}_{x_i=w} \alpha_i,
\end{align*}
where $\mathbb{I}$ is an indicator function. We omitted the conditioning variables $\hat{y}_{<t}$ and $X$ for brevity.

The mixture coefficient $m_t$ indicates the degree to which the decoder generates the next symbol from its own vocabulary. When $m_t$ is close to $1$, the decoder is free to {\em generate} any symbol from the vocabulary regardless of its existence in the input. On the other hand, when $m_t$ is close to $0$, the decoder ignores $p_{\text{gen}}$ and {\em copies} over one of the input symbols to the output using $p_{\text{copy}}$. 

\paragraph{Mismatch between training and decoding}

~\citet{see2017get} observed that the statistics of the mixture coefficient $m_t$ differs significantly between training and testing. During training, the average of the mixture coefficients was found to converge to around 0.53, while it was much smaller, close to 0.17, during testing when summaries generated from the trained model (i.e. without teacher forcing). Furthermore, most of the generated $n$-grams turn out to be exact copies from the input paragraph with this model. Our analysis also corroborates this observation. 

\subsection{Copy-Controlled Decoding}

With a conditional neural language model, such as the neural abstractive summarization system here, we often use beam search to find a target sequence. At each time step, beam search collects top-$K$ prefixes according to a scoring function 
defined as
\[
s(y_{\leq t}, X) = \log p(y_{\leq t}| X) = \sum_{t'=1}^t \log p(y_t'|y_{<t'}, X).
\]

In order to bridge between training and decoding in terms of the amount of copying, we propose a new scoring function: 
\begin{align}
\label{eq:copy-control}
s(y_{\leq t}, X) = \sum_{t'=1}^t & \log p(y_{t'} | y_{< t}, X)  \\
&- \eta_t
\max(0, m^* - \bar{m}_{t'}),
\nonumber
\end{align}
where $m^*$ is a target coefficient, $\eta_t$ is a time-varying penalty strength, and 
\mbox{$\bar{m}_{t'} = \frac{1}{t'}\sum_{t''=1}^{t'} m_{t''}$}. We use the following scheduling of $\eta_t$ to ensure the diversity of hypotheses in the early stage of decoding:
$\eta_t = t \cdot \eta_0,$ although other scheduling strategies should explored in the future.

The penalty term in Eq.~\eqref{eq:copy-control} allows us to softly eliminate any hypothesis whose average mixture coefficients thus far has been too far away from the intended ratio. The target average may be selected via validation or determined manually. 
Later in the experiments, we present both quantitative and qualitative impacts of the proposed copy-controlled decoding procedure.

\begin{figure}[t]
\small
\centering

\begin{minipage}{0.1\columnwidth}
\flushright
\rotatebox{90}{(a) -coverage penalty}
\end{minipage}
\hfill
\begin{minipage}{0.89\columnwidth}
\includegraphics[width=0.8\columnwidth,clip=True,trim=0 50 0 0]{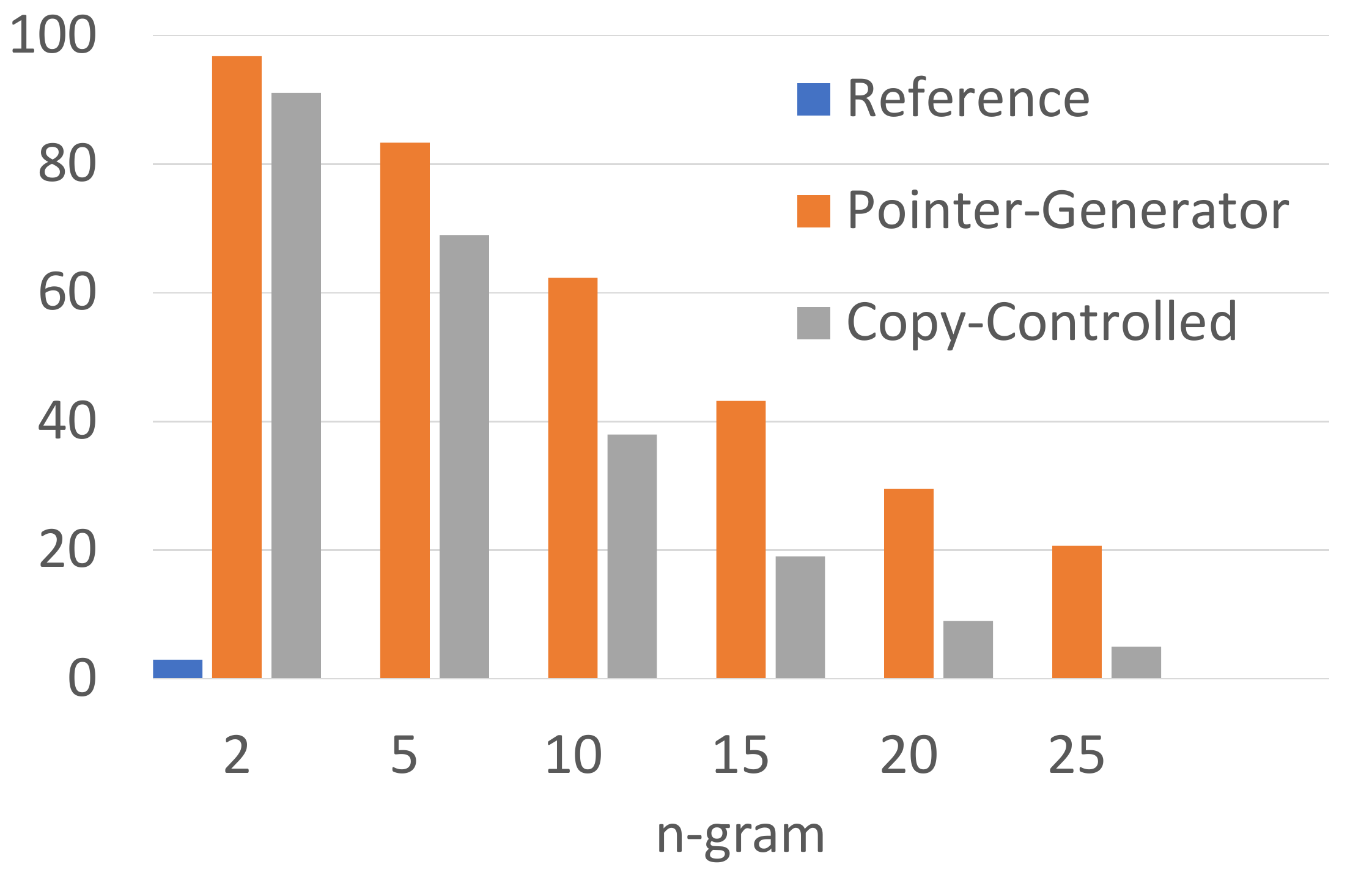}
\end{minipage}

\begin{minipage}{0.1\columnwidth}
\flushright
\rotatebox{90}{(b) +coverage penalty}
\end{minipage}
\hfill
\begin{minipage}{0.89\columnwidth}
\includegraphics[width=0.8\columnwidth]{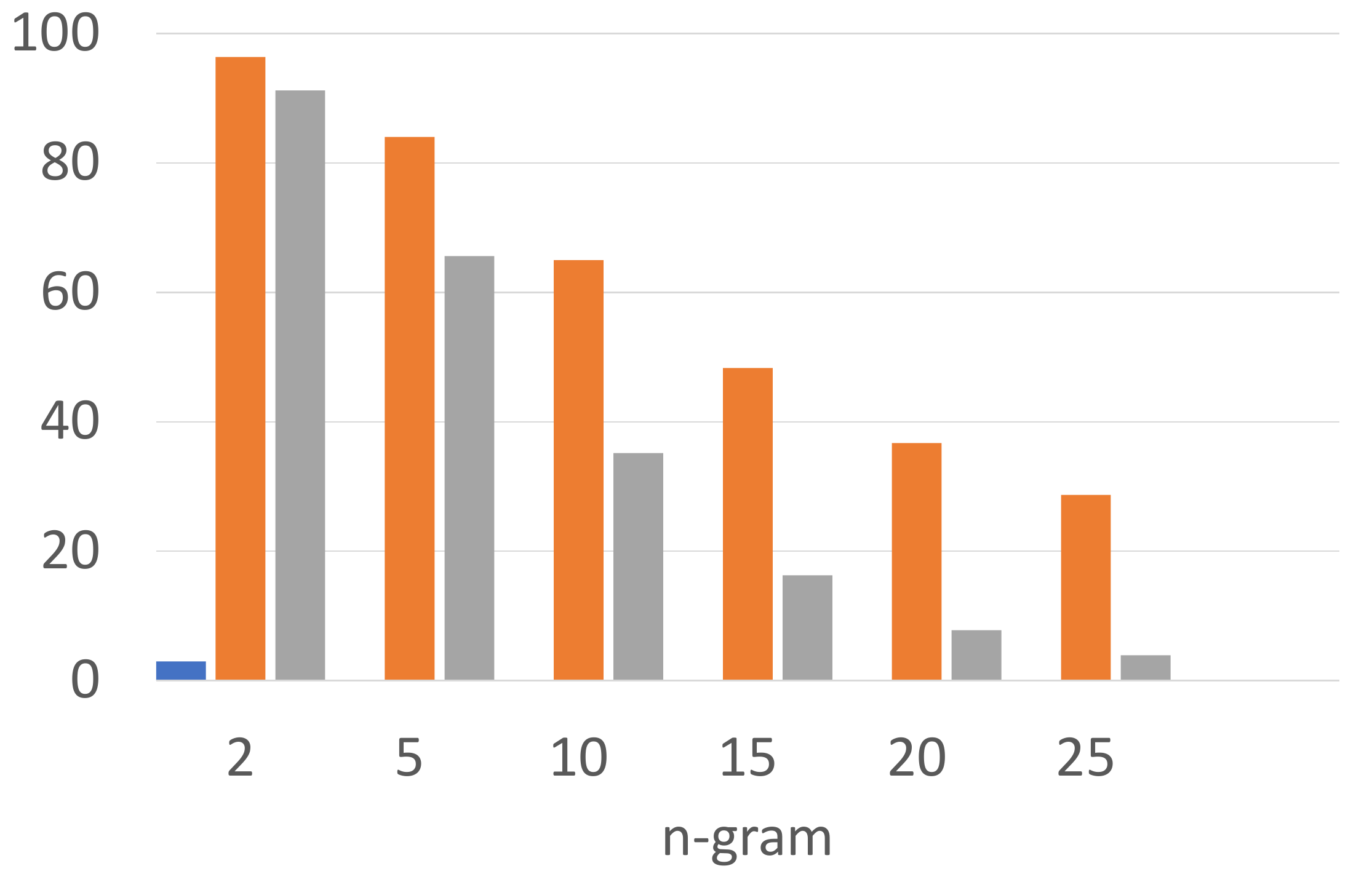}
\end{minipage}

\vspace{-3mm}
\caption{
\label{fig:ngram-overlap}
The $n$-gram overlap (\%) (a) with and (b) without the coverage penalty. We observe reduction with the proposed copy-controlled decoding.
}
\vspace{-1.5mm}
\end{figure}

\begin{table}[t]
\centering
\small
\begin{tabular}{l || l l | l l}
& \multicolumn{2}{c|}{} & \multicolumn{2}{c}{+coverage} \\
& Std. & C-C & Std. & C-C \\
\toprule
ROUGE-1 & 35.77 & 34.10 & 38.82 & 36.22 \\
ROUGE-2 & 15.28 & 14.02 & 16.81 & 14.69 \\
ROUGE-L & 32.54 & 31.15 & 35.71 & 33.60 \\
\midrule
METEOR & 15.03 & 13.73 & 16.73 & 14.57 \\
METEOR$^\star$ & 16.33 & 14.96 & 18.14 & 15.84 \\
\midrule
\# of tokens & 56.10 & 50.47 & 59.87 & 50.30
\end{tabular}
\caption{
\label{tab:quality}
The summary quality measured in ROUGE-X and METEOR. Standard: Standard abstractive beam search for decoding. C-C: copy-controlled decoding. METEOR$^\star$: METEOR +stem +synonym +paraphrase.}
\vspace{-6.5mm}
\end{table}

\paragraph{Related Work}

Although decoding algorithms for conditional neural language models have received relatively little interest, there have been a few papers that have aimed at improving existing decoding algorithms. One line of such research has been on augmenting the score function of beam search. For instance,~\citet{li2016simple} proposed a diversity promoting term, and shortly after,~\citet{li2017learning} generalized this notion by learning such a term. Another line of research looked at replacing manually-designed decoding algorithms with trainable ones based on a recurrent network~\citep{gu2016learning,gu2017trainable}. The proposed copy-controlled decoding falls in the first category and is applicable to any neural generation model that partly relies on the copy mechanism, such as neural machine translation~\citep{gulcehre2016pointing,gu2017search} and data-to-document generation~\citep{wiseman2017challenges}. 

\begin{table*}[th!]
\centering
\small
\begin{tabular}{|p{\linewidth}|}
\hline
\textbf{Article}: a federal judge has ordered the defense department to release photos that allegedly show detainees being abused in detention centers in iraq and afghanistan during the bush administration . the photos wo n't be made public right away . in an order issued friday , u.s. district judge alvin k. hellerstein of the southern district of new york granted the government 60 days to appeal . the aclu sued the defense department in 2003 to have the photos made public . it 's not clear how many photos are involved or where the pictures were taken (...) \\
\hline
\textbf{Standard + coverage}:  \colorbox{blue!6}{u.s.} \colorbox{blue!2}{district} \colorbox{blue!3}{judge} \colorbox{blue!5}{alvin} \colorbox{blue!0}{k.} \colorbox{blue!0}{hellerstein} \colorbox{blue!20}{of} \colorbox{blue!2}{the} \colorbox{blue!0}{southern} \colorbox{blue!0}{district} \colorbox{blue!5}{of} \colorbox{blue!0}{new} \colorbox{blue!0}{york} \colorbox{blue!16}{granted} \colorbox{blue!4}{the} \colorbox{blue!1}{government} \colorbox{blue!6}{60} \colorbox{blue!1}{days} \colorbox{blue!4}{to} \colorbox{blue!2}{appeal} \colorbox{blue!25}{.} \colorbox{blue!7}{the} \colorbox{blue!1}{aclu} \colorbox{blue!17}{sued} \colorbox{blue!3}{the} \colorbox{blue!0}{defense} \colorbox{blue!1}{department} \colorbox{blue!4}{in} \colorbox{blue!0}{2003} \colorbox{blue!9}{to} \colorbox{blue!2}{have} \colorbox{blue!1}{the} \colorbox{blue!0}{photos} \colorbox{blue!4}{made} \colorbox{blue!1}{public} \colorbox{blue!24}{.} \colorbox{blue!11}{it} \colorbox{blue!12}{'s} \colorbox{blue!8}{not} \colorbox{blue!2}{clear} \colorbox{blue!4}{how} \colorbox{blue!1}{many} \colorbox{blue!2}{photos} \colorbox{blue!1}{are} \colorbox{blue!0}{involved} \colorbox{blue!3}{or} \colorbox{blue!1}{where} \colorbox{blue!1}{the} \colorbox{blue!2}{pictures} \colorbox{blue!5}{were} \colorbox{blue!1}{taken} \colorbox{blue!19}{.} \\
\hline
\textbf{C-C + coverage}: \colorbox{blue!7.5}{federal} \colorbox{blue!3.7}{judge} \colorbox{blue!19.9}{orders} \colorbox{blue!8.0}{defense} \colorbox{blue!6.9}{department} \colorbox{blue!16.2}{to} \colorbox{blue!7.7}{release} \colorbox{blue!12.5}{photos} \colorbox{blue!18.8}{of} \colorbox{blue!2.4}{detainees} \colorbox{blue!14.6}{being} \colorbox{blue!5.3}{abused} \colorbox{blue!16.6}{in} \colorbox{blue!1.9}{afghanistan} \colorbox{blue!16.5}{and} \colorbox{blue!1.1}{iraq} \colorbox{blue!24.7}{.} \colorbox{blue!5.3}{photos} \colorbox{blue!17.3}{may} \colorbox{blue!14.9}{have} \colorbox{blue!10.1}{been} \colorbox{blue!12.2}{made} \colorbox{blue!8.1}{public} \colorbox{blue!9.3}{right} \colorbox{blue!4.6}{away} \colorbox{blue!30.0}{,} \colorbox{blue!8.5}{judge} \colorbox{blue!24.7}{says} \colorbox{blue!29.6}{.} \colorbox{blue!12.1}{aclu} \colorbox{blue!26.3}{says} \colorbox{blue!20.0}{photos} \colorbox{blue!19.6}{are} \colorbox{blue!12.0}{``} \colorbox{blue!0.7}{the} \colorbox{blue!0.4}{best} \colorbox{blue!0.3}{evidence} \colorbox{blue!6.4}{of} \colorbox{blue!1.7}{what} \colorbox{blue!2.0}{took} \colorbox{blue!0.9}{place} \colorbox{blue!4.6}{in} \colorbox{blue!0.6}{the} \colorbox{blue!0.5}{military} \colorbox{blue!2.4}{'s} \colorbox{blue!1.1}{detention} \colorbox{blue!3.3}{centers} \colorbox{blue!29.5}{''}  \\
\hline
\hline
\textbf{Article}: a pennsylvania community is pulling together to search for an eighth-grade student who has been missing since wednesday . the search has drawn hundreds of volunteers on foot and online . the parents of cayman naib , 13 , have been communicating through the facebook group `` find cayman '' since a day after his disappearance , according to close friend david binswanger . newtown police say cayman was last seen wearing a gray down winter jacket , black ski pants and hiking boots . he could be in the radnor-wayne area , roughly 20 miles from philadelphia (...) \\
\hline
\textbf{Standard + coverage}: \colorbox{blue!6}{the} \colorbox{blue!3}{search} \colorbox{blue!10}{has} \colorbox{blue!4}{drawn} \colorbox{blue!4}{hundreds} \colorbox{blue!2}{of} \colorbox{blue!1}{volunteers} \colorbox{blue!3}{on} \colorbox{blue!0}{foot} \colorbox{blue!3}{and} \colorbox{blue!0}{online} \colorbox{blue!28}{.} \colorbox{blue!4}{the} \colorbox{blue!0}{parents} \colorbox{blue!4}{of} \colorbox{blue!0}{cayman} \colorbox{blue!0}{naib} \colorbox{blue!9}{,} \colorbox{blue!0}{13} \colorbox{blue!20}{,} \colorbox{blue!7}{have} \colorbox{blue!2}{been} \colorbox{blue!1}{communicating} \colorbox{blue!2}{through} \colorbox{blue!1}{the} \colorbox{blue!0}{facebook} \colorbox{blue!4}{group} \colorbox{blue!6}{``} \colorbox{blue!0}{find} \colorbox{blue!0}{cayman} \colorbox{blue!13}{''} \colorbox{blue!4}{since} \colorbox{blue!2}{a} \colorbox{blue!0}{day} \colorbox{blue!6}{after} \colorbox{blue!1}{his} \colorbox{blue!0}{disappearance} \colorbox{blue!20}{.} \\
\hline
\textbf{C-C + coverage}: \colorbox{blue!6.8}{cayman} \colorbox{blue!4.5}{naib} \colorbox{blue!17.4}{,} \colorbox{blue!1.1}{13} \colorbox{blue!28.2}{,} \colorbox{blue!12.9}{has} \colorbox{blue!18.2}{been} \colorbox{blue!3.6}{missing} \colorbox{blue!17.6}{since} \colorbox{blue!2.5}{wednesday} \colorbox{blue!30.0}{.} \colorbox{blue!7.4}{the} \colorbox{blue!1.6}{search} \colorbox{blue!14.6}{has} \colorbox{blue!7.2}{drawn} \colorbox{blue!6.3}{hundreds} \colorbox{blue!3.0}{of} \colorbox{blue!1.6}{volunteers} \colorbox{blue!4.4}{on} \colorbox{blue!0.9}{foot} \colorbox{blue!6.4}{and} \colorbox{blue!1.0}{online} \colorbox{blue!30.0}{.} \colorbox{blue!8.8}{he} \colorbox{blue!12.0}{could} \colorbox{blue!7.1}{be} \colorbox{blue!2.6}{in} \colorbox{blue!1.6}{the} \colorbox{blue!1.2}{radnor-wayne} \colorbox{blue!1.0}{area} \colorbox{blue!9.1}{,} \colorbox{blue!1.3}{roughly} \colorbox{blue!0.3}{20} \colorbox{blue!0.2}{miles} \colorbox{blue!1.9}{from} \colorbox{blue!0.4}{philadelphia} \colorbox{blue!19.9}{.}  \\
\hline
\end{tabular}

\vspace{-1mm}
\caption{
\label{tab:outputs}
Sample summaries generated by the vanilla beam search and the proposed copy-controlled decoding. Colors indicate the strengths of the mixture coefficients. Darker shade of blue indicates high mixture coefficient value i.e., more generation, and lighter color indicates low value i.e. more copying.}
\vspace{-2mm}
\end{table*}

\section{Experiments}

We use the pretrained pointer-generator network, trained on CNN/DailyMail data~\citep{hermann2015teaching,nallapati2016summarunner}, provided by \citet{see2017get}.

The pretrained network is provided together with the code based on which we implement the proposed copy-controlled decoding. It should be noted that our work is not strictly comparable to the abstractive work done by~\citet{Nallapati2016AbstractiveTS} as the latter was trained and evaluated on the anonymized dataset and used pretrained word embeddings. 

\subsection{Quantitative Analysis}


\paragraph{Controlling $m_t$}

We first test whether the proposed scoring function in Eq.~\eqref{eq:copy-control} does indeed allow us to control the mixture coefficient. When forced to generate summaries of a randomly drawn subset of 500 validation examples, the pointer-generator network with the original scoring function resulted in the average mixture coefficients of 0.24 and 0.26 respectively with and without the coverage penalty. With the target mixture coefficient $m^*$ set to 0.4 and the penalty coefficient $\eta=0.5$, we observed that the average mixture coefficient increased to 0.29 and 0.33, respectively, with and without the coverage penalty. While the mixture coefficients increased on average, the ROUGE-1 scores roughly stayed at the same level, going from 27.49 and 28.87 to 27.63 and 28.86 (w/ and w/o the coverage penalty). We observed the similar trends with other evaluation metrics.
Based on this observation, we use $m^*=0.4$ and $\eta=0.5$ with the full test set from here on.

\paragraph{Abstractiveness}

We then investigate the overlap between the input paragraph and summary to measure the novelty in the generated summary. We count the number of $n$-grams in a summary and those that also occur in the original article and look at the ratio (\%). To draw a clear picture, we do so for a wide range from $n=2$ up to $n=25$. We report the number of $n$-gram overlaps in Fig.~\ref{fig:ngram-overlap}.

The first observation we make is that there is almost no overlap between the input paragraph and its reference summary. There are a few overlapping bi-grams, $\leq 2$ on average. On the other hand, the summaries generated by the pointer-generator network exhibit significantly more $n$-gram overlap. For instance, over 20\% 25-grams are found exactly as they are in the input paragraph on average. Furthermore, the overlap increases when the coverage penalty is used in decoding, suggesting that its success may not be only due to the removal of repeated sentences but also due to even more aggressive copying of phrases/sentences from the input paragraph. 
We observe that the proposed copy-controlled decoding algorithm effectively reduces the $n$-gram overlap. We observe the significant reduction of overlaps especially when $n$ is large. For instance, when the coverage penalty was used, the 25-gram overlap decreased from 28.72\% to 3.93\%, which is quite significant.  


\paragraph{Summary Quality}

Along with significant reduction in the overlap between the generated summary and input, we also observed slight degradation in the summary quality measured in terms of ROUGE and METEOR scores, as shown in Table~\ref{tab:quality}. 
The small drop in these scores is however not discouraging, because the proposed copy-controlled (C-C) decoding generally generates slightly shorter summaries, while the recall-based ROUGE (or METEOR) score prefers longer summaries.
This suggests the effectiveness of our approach, since ROUGE (or METEOR) does not take into account the length of a hypothesis but only considers the recall rate of $n$-grams. 
Finally, we note that the ROUGE and METEOR scores of the summaries generated by the pointer generator network, released by ~\citet{see2017get} is lower than what they have reported. We believe this neither contradicts nor discounts our contribution, as the proposed decoding works on top of any pre-trained summarizer which relies on beam search during inference.

\subsection{Qualitative Analysis}

We illustrate the benefits of the copy-controlled decoding with examples 
in Table~\ref{tab:outputs}. The heatmap of the mixture co-efficients show that copy-controlled {\em generates} more often (shown by darker shade) than the standard decoding model. Standard decoding is fully extractive, copying over full sentences from the input. Whereas, copy-controlled decoding is more abstractive. It generates new words "orders" and "says" that are not part of the input. Also, it turns out that favoring higher mixture co-efficients improves the ability to condense information. In both examples, controlled-decoding condenses information present in two different sentences to generate a single output sentence. 

We conjecture that an occasionally high mixture coefficient, encouraged by the proposed copy-controlled decoding, disrupts the sequence of copy operations, enabling the attention mechanism to jump to another part of the input paragraph. This leads to a more compact summary that compresses information from multiple sentences distributed across the input paragraph. We leave more in-depth analysis for future work.

\section{Conclusion}

In this paper, we confirmed that a recently popular neural abstractive summarization approach largely performs extractive summarization when equipped with the copy mechanism.  To address this, we proposed a copy-controlled decoding procedure that introduces a penalty term to the scoring function used during beam search and empirically validated its effectiveness. Our proposed mechanism currently only modifies the decoding behavior. A future direction would be to investigate ways to enforce abstractiveness of a summary during training.

\section{Acknowledgement}
We would like to thank the authors of \citet{see2017get} for their publicly available, well documented code. Noah Weber and Niranjan Balasubramanian were supported in part by the National Science Foundation under Grant IIS-1617969. Kyunghyun Cho was partly supported by Samsung Advanced Institute of Technology (Next Generation Deep Learning: from pattern recognition to AI) and Samsung Electronics (Improving Deep Learning using Latent Structure).

\bibliographystyle{acl_natbib}
\bibliography{naaclhlt2018}

\end{document}